\documentclass[twocolumn,letterpaper]{IEEEAerospaceCLS}

\usepackage[]{graphicx,float,latexsym,amssymb,amsfonts,amsmath,amstext,times,cite,nomencl}

\newcommand{\ignore}[1]{}
\makenomenclature

\begin{document}
\title{End-to-End Imitation Learning for Optimal Asteroid Proximity Operations}

\author{
Patrick Quinn\\ 
Florida Institute of Technology\\
150 W. University Blvd.\\
Melbourne, FL 32901\\
pquinn2019@my.fit.edu
\and 
George Nehma\\ 
Florida Institute of Technology\\
150 W. University Blvd.\\
Melbourne, FL 32901\\
gnehma2020@my.fit.edu
\and 
Dr. Madhur Tiwari\\ 
Florida Institute of Technology\\
150 W. University Blvd.\\
Melbourne, FL 32901\\
mtiwari@fit.edu
\thanks{\footnotesize 979-8-3503-5597-0/25/$\$31.00$ \copyright2025 IEEE}
}

\maketitle

\thispagestyle{plain}
\pagestyle{plain}

\maketitle

\thispagestyle{plain}
\pagestyle{plain}

\begin{abstract}
Controlling spacecraft near asteroids in deep space comes with many challenges. The delays involved necessitate heavy usage of limited onboard computation resources while fuel efficiency remains a priority to support the long loiter times needed for gathering data. Additionally, the difficulty of state determination due to the lack of traditional reference systems requires a guidance, navigation, and control (GNC) pipeline that ideally is both computationally and fuel-efficient, and that incorporates a robust state determination system. In this paper, we propose an end-to-end algorithm utilizing neural networks to generate near-optimal control commands from raw sensor data, as well as a hybrid model predictive control (MPC) guided imitation learning controller delivering improvements in computational efficiency over a traditional MPC controller.
\end{abstract} 

\tableofcontents

\nomenclature{CNN\quad\enspace}{Convolutional Neural Network.}
\nomenclature{MPC\quad\enspace}{Model Predictive Controller.}
\nomenclature{LSTM\quad}{Long Short-Term Memory.}
\nomenclature{MLP\quad\enspace}{Multi-Layer Perceptron.}
\nomenclature{ROS 2\quad}{Robot Operating System 2.}

\printnomenclature

\section{Introduction}

As demand for the exploration of asteroids continues to increase, so too have the demands for satellite performance near these asteroids. Long communication delays with Earth, low availability of onboard computational resources, need for efficient fuel usage on long missions, and lack of knowledge of the operational environment's exact characteristics all pose a daunting challenge for the design of GNC systems within these satellites. With the rapid proliferation of learning based techniques in robotics \cite{NewRef1,NewRef2,NewRef3,NewRef4,NewRef5,NewRef6,NewRef7,NewRef8,NewRef9,NewRef10,NewRef111}, new solutions have emerged. To fulfill these demanding requirements, we propose a control policy which utilizes extensive ground-based computation resources to pre-train a robust model \cite{Li2023-if} which is both computationally efficient and fuel efficient\cite{Kalaria2023-su,Scaramuzza2022-le}. This is accomplished with the usage of imitation learning, a previously developed MPC \cite{Tiwari2022-jt, Tiwari2022-ki}, and basic assumptions about the characteristics of the target asteroid.

Imitation learning leverages offline computation and simulation to train models with the goal of replicating the behavior of an expert agent, typically in the form of a human or a traditional control pipeline \cite{Attia2018-cy,Zare2023-kf}. This process allows controllers to be developed which can be deployed in situations with uncertain dynamics as the controller no longer relies on exact knowledge of system dynamics, and allows for the development of controllers that offer human-like performance in situations where human interaction with the system is otherwise difficult or impractical\cite{Lee2018-qh,Pan2020-rt}. These models can further be trained to use otherwise abstract data such as raw sensor data as inputs, allowing for the elimination of state estimation from the GNC pipeline \cite{Chen2020-ab,Pan2020-rt,Scaramuzza2022-le}. This may be achieved by giving the expert agent access to full knowledge of current system states as well as access to the system dynamics, while the trainee may only have access to basic sensor readings as input \cite{Zhang2016-ip,Chen2020-ab}. Our implementation is heavily influenced by the ``Learning by Cheating'' \cite{Chen2020-ab} approach, where the expert agent has access to privileged information on its exact state and the system dynamics. This allows an optimal controller such as a MPC to be used in the generation of expert demonstrations rather than a human or more ``brute force'' style traditional control methods.

When deployed in real environments, imitation learning based controllers are often observed to encounter difficulties with generalizability as deviations from the states they were trained on typically causes model performance to degrade, leading to further deviations from the set of states the model was trained on and even worse performance \cite{Alexander2024-my,Pan2020-rt}. This is known as the covariate shift problem. As it often results in a complete failure of the control system, in this work we attempt to minimize the issues associated with covariate shift by including recovery trajectories in the training dataset \cite{Alexander2024-my,Codevilla2018-sq}. By purposefully introducing trajectory deviations into the expert dataset which the MPC can easily recover from, we provide useful information to the trained controller on how to react to its own deviations from the optimal trajectory when deployed.

Our algorithm is trained to imitate the performance of the previously mentioned MPC using only raw lidar data and a user-selected desired final state as input. We aim to improve the speed and computational consistency of generating new control commands over the MPC while maintaining near optimal control outputs and the ability to navigate in close proximity to an asteroid. We then went on to develop a hybrid controller for when system states are occasionally available which refers to an MPC to validate performance of the imitation learning based controller as a form of runtime assurance. The resulting hybrid MPC guided imitation learning controller delivered consistently stable control while significantly reducing power usage and increasing average inference speed. 

\section{Methodology}

In order to develop our model, the first step was to gather a large and diverse dataset characterizing the MPC's control response within the operational regime we planned to deploy our final model in. This was accomplished by collecting data on simulations of approximately 400 randomly selected point-to-point transits in close proximity to the asteroid Kleopatra, as performed by the MPC. In the first 100 transits, the MPC's control outputs were applied directly to the satellite without modification. As expected, these transits were highly optimal and as a result while providing a great performance offered little in the way of thrust commands for the model to learn from. Data demonstrating actually getting on an optimal trajectory, as well as mitigating the effects of errors or other disturbances proved to be far more important and as such the other approximately 300 trajectories collected attempted to address this. For these trajectories, after the MPC generated a control output a disturbance was added to the control output before it was applied to the satellite.  This disturbance was reset every 100 seconds and had a randomized limit on its intensity, allowing it to range from having zero effect on the satellite to having twice the magnitude of the MPC's requested output, allowing the satellite to potentially thrust completely opposite to the optimal direction. This disturbance was updated incrementally between time steps from a small normal distribution with the aim that it would evolve in a slow random walk rather than acting as random noise. This was done to more realistically simulate the errors that the trained controller would exhibit as it would be using a deterministic model. In the cases where the disturbance was applied, limitations were placed such that physical thruster constraints were still respected after the application of the disturbance. Data was collected on the current satellite state, commanded final state (the targeted endpoint of each transit), and the thrust command generated by the MPC at each time step. The simulation included estimations of forces from a polyhedral gravity model and was performed in MATLAB.

\begin{figure}[ht]
    \centering
    \includegraphics[width=2.15in]{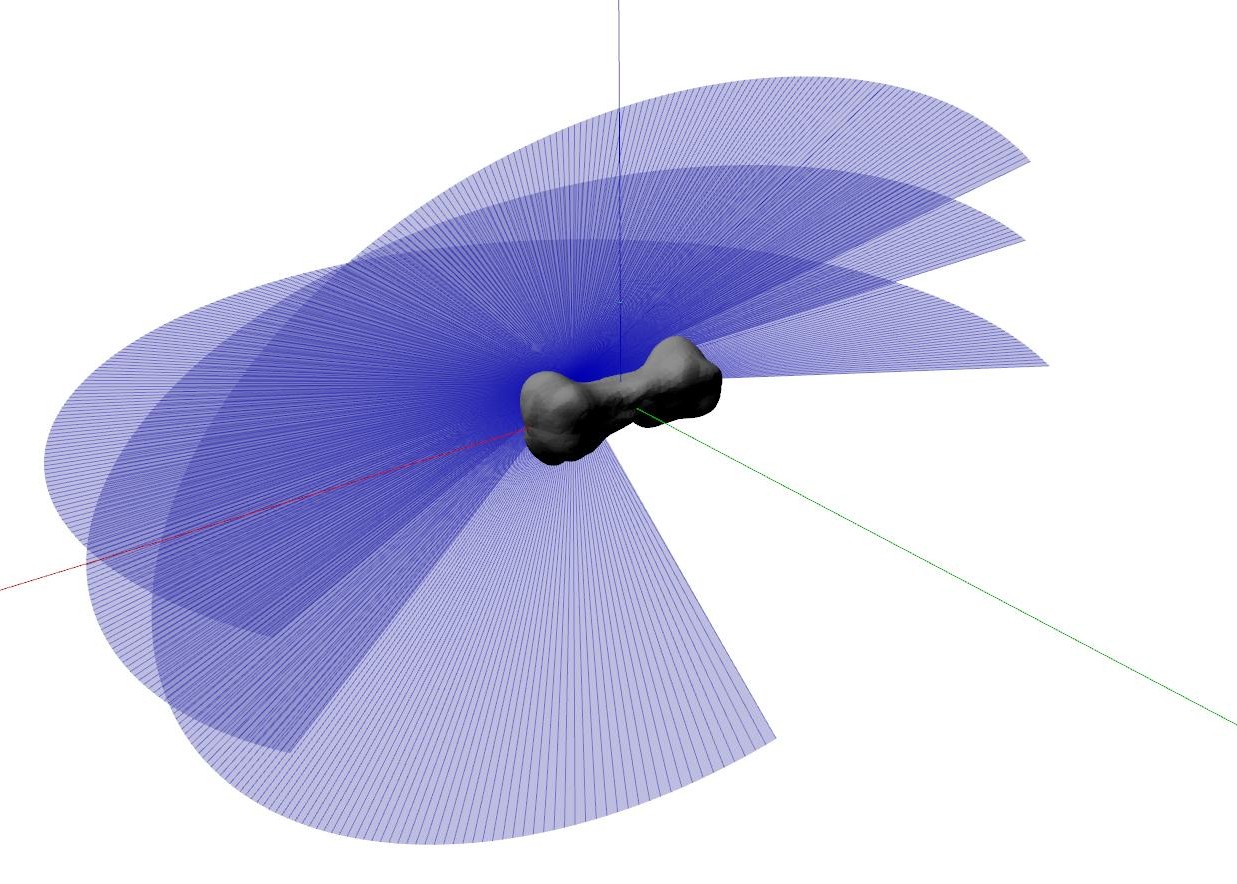}
    \caption{Visualization of 3 of the 12 lidar sensors and Kleopatra in Gazebo}
    \label{fig:SimScreenshot}
\end{figure}

\begin{figure}[ht]
    \centering
    \includegraphics[width=3.25in]{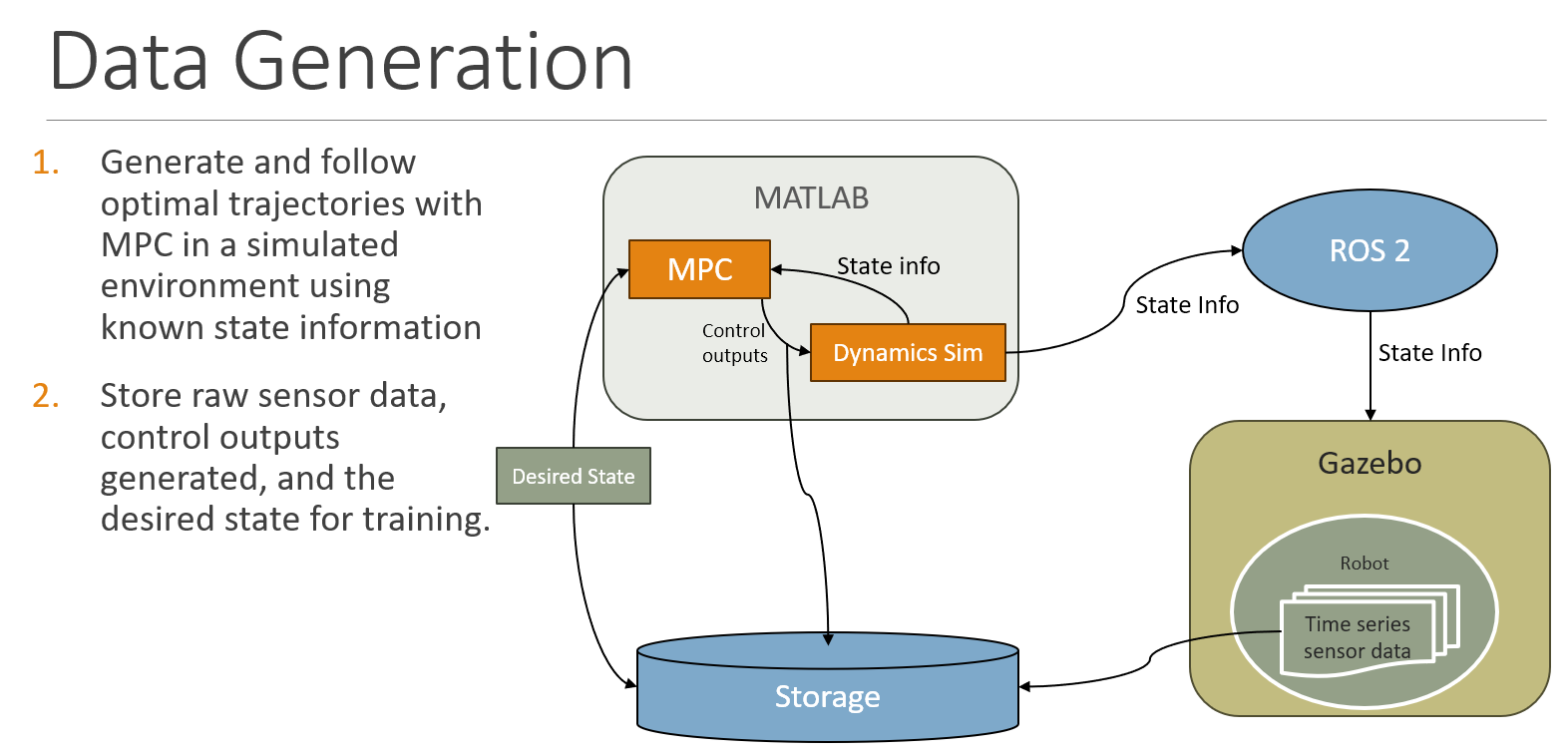}
    \caption{Diagram detailing training dataset generation}
    \label{fig:DataGeneration}
\end{figure}

Next, realistic sensor data had to be gathered for each recorded satellite state. This sensor data is what would later be used in imitation learning so sensory inputs could be associated with expert control outputs by the trained model. This was accomplished by utilizing a MATLAB script, ROS 2, and Gazebo to simulate a spherical lidar array's returns along each trajectory flown by the satellite while using the MPC. The array used 12 planar lidar sensors, each rotated an equal amount around a common axis, to produce lidar ``frames'' with a resolution of 360x12. A visualization of the first three of the sensors in the asteroid environment can be seen in Figure~\ref{fig:SimScreenshot}. Lidar scan rays which didn't intersect with the 3D model for Kleopatra used in the Gazebo simulation returned values of infinity as a range reading. While this makes intuitive sense, values of infinity had to be processed to be a constant negative value (chosen arbitrarily as -100) before being fed to the trained model so that computations could be performed on these returns by the model. A constant negative value was chosen as the lidar sensors would otherwise never give such a result in normal operation, so these values could effectively be associated by the model to mean the lidar didn't ``see'' anything in that particular direction. These first two steps are summarized in Figure~\ref{fig:DataGeneration}. 

\begin{figure}[ht]
    \centering
    \includegraphics[width=2.2in]{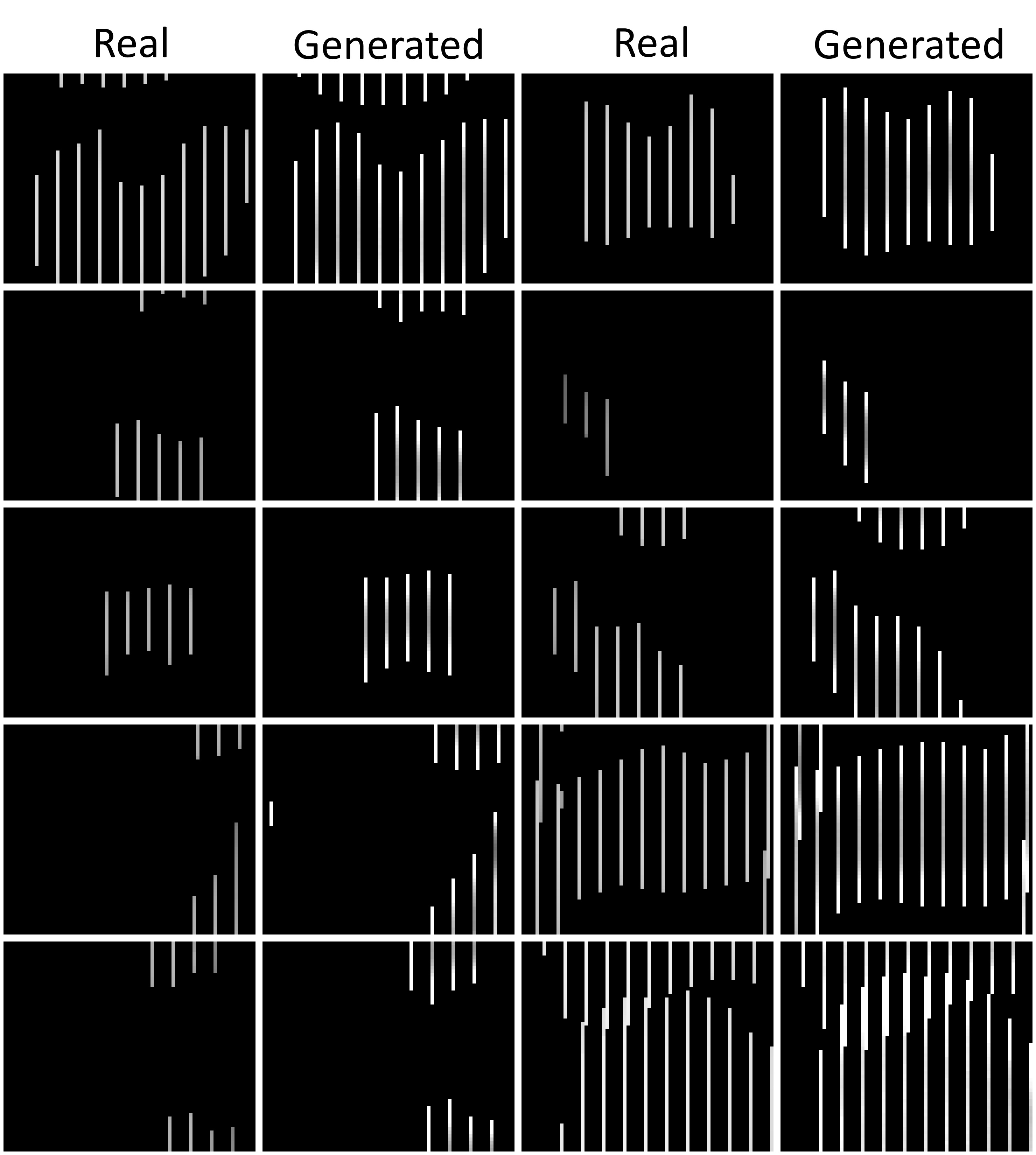}
    \caption{Comparisons between real lidar readings and generated readings.}
    \label{fig:ComparisonImage}
\end{figure}

In order to improve generalized model performance as well as to speed up the collection of lidar data for trajectories, synthetic dataset generation was employed to generate large amounts of lidar data without having to use the relatively slow Gazebo, ROS 2, and MATLAB simulation pipeline. By adding in synthetic data which will inevitably have some unpredictable errors due to the imperfect nature of the model when compared to the data generated in the Gazebo simulation, we hope to better prepare the trained controller for any variance between the simulated environment and the real environment when deployed. This approach was used on the $\sim$300 trajectories where disturbances were applied and relied on an MLP trained to output the lidar return expected from the satellite for any given state.  A set of comparison images between captured lidar readings and generated lidar readings (each scaled from 360x12 to 60x72 for convenience of visualization) can be seen in Figure~\ref{fig:ComparisonImage}. In order to train this model, lidar data was collected from Gazebo in a grid pattern of positions near Kleopatra chosen to be as large as possible while still remaining feasible in terms of time taken to collect data. This ended up capturing lidar data from a 340 km cube centered on Kleopatra, leaving a wide margin for data generation on trajectories which moved further away from Kleopatra.

\begin{figure}[ht]
    \centering
    \includegraphics[width=3.25in]{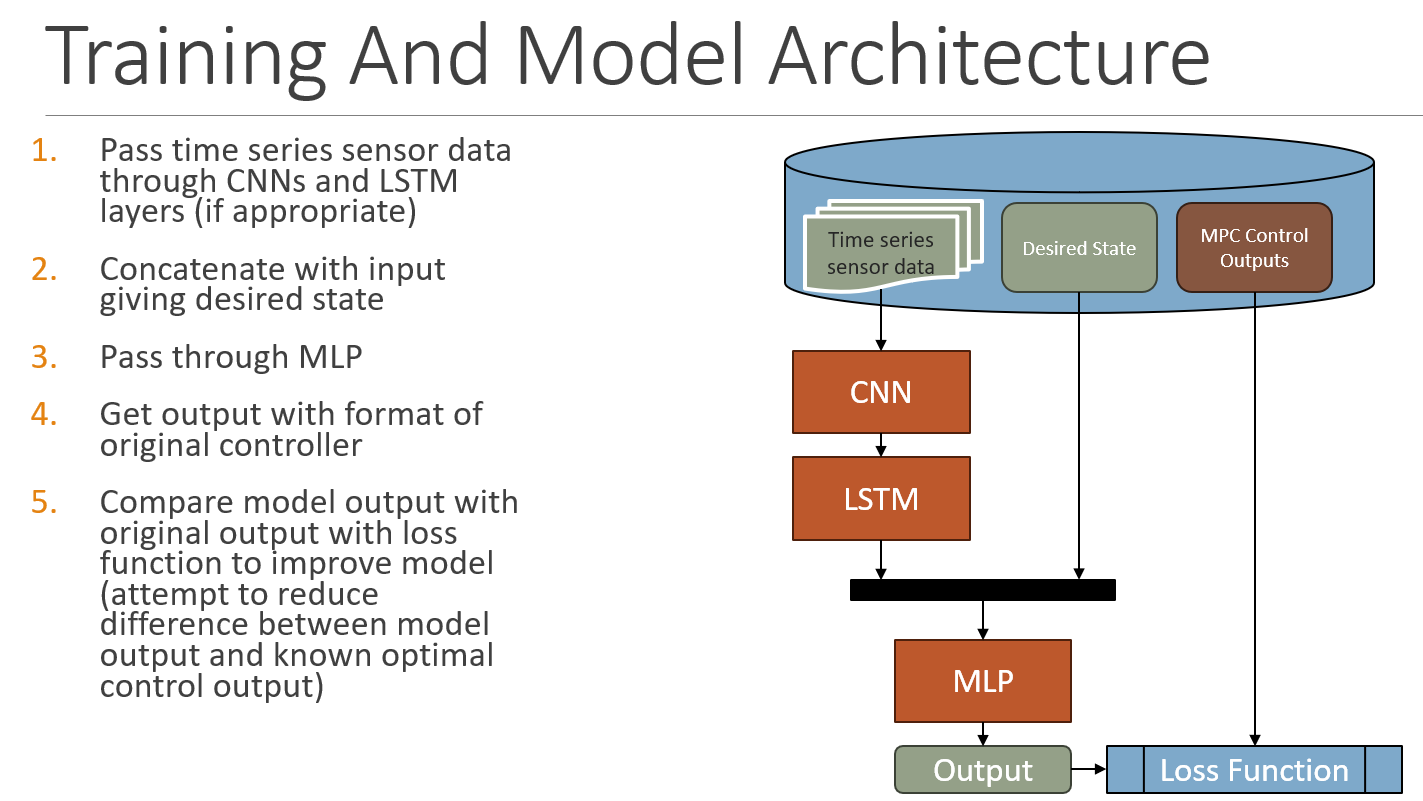}
    \caption{Diagram detailing training and model architecture}
    \label{fig:TrainingArchitecture}
\end{figure}

Finally, model training could begin. The model architecture chosen used multiple layers of CNNs and LSTM networks to process the time series lidar data being fed to the model. Next, this processed sensor data was concatenated with the desired final state and fed through an MLP\cite{Kaufmann2020-mr}. The model then outputs values in the same format as the MPC control output (a force command along each axis). The difference between the neural network's generated force command and the force commanded by the MPC was then compared by a root mean square error (RMSE) loss function. Model weights were adjusted incrementally through the training process to minimize the difference observed between the control commands generated by the MPC and the neural network. The model with the best performance on a validation dataset consisting of 20\% of the original data was then selected as a final candidate for deployment. This trained model was tested on a separate test dataset containing 20\% more of the original data for an initial evaluation of model performance. The remaining 60\% of the dataset was used as training data. This training process is detailed in Figure~\ref{fig:TrainingArchitecture}. 

\begin{figure*}
\centering
\includegraphics[width=6in]{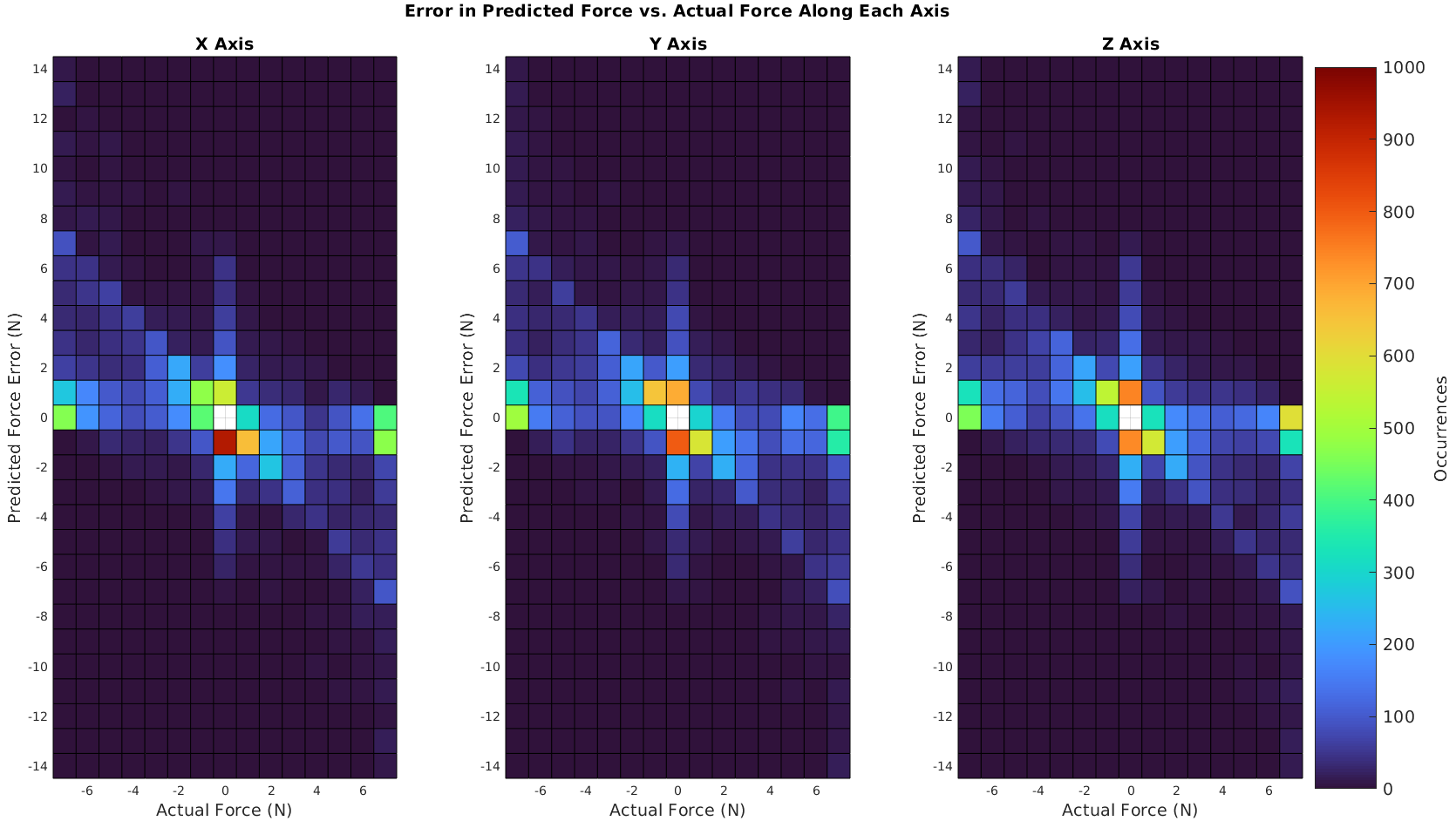}
\caption{2D histograms of force commands generated along all axes for MPC (actual force) and imitation learning model (predicted force) error compared to MPC output when using the test dataset after model training.}
\label{fig:FullFinalHist}
\end{figure*}

Once the model was trained, initial performance evaluations were made based on an analysis of the correlation in responses between the MPC and the trained agent when fed the test dataset. Since we aim to retain the near optimal performance of the MPC, the hope is to see a perfect 1:1 correlation between the force commanded along an axis by the MPC and the force commanded along that axis by the trained model. In Figure~\ref{fig:FullFinalHist}, this correlation is represented with 2D histograms for each thrust axis. To highlight the relevant data for this analysis, the axes have been scaled, obscuring some data (represented by the white square in the center of the histograms). This was done as the vast majority of the time, both the MPC and trained model were in agreement that no thrust was needed. This fact isn't useful to the improvement of the model's performance, so instead the color scale of the histograms were limited to 1,000 occurrences, which enabled the more important results with non-zero values to be better visualized. The trained model seemed to perform well, with a response generally in agreement with the MPC. Notably, there were a non-insignificant number of cases where the trained model called for a 0 N thrust output in an axis while the MPC was calling for a full ($\pm$ 7 N) thrust output. Additionally, when the MPC called for a 0 N thrust output along an axis, the trained model was seen to often return some small thrust value. This is likely a result of the inclusion of recovery trajectories in the training data set, where these small corrective maneuvers were common and (at least to the model) seemingly random, as they were a response to the random disturbance employed in the training dataset, which the model was not provided data on.

\begin{figure}[ht]
\centering
\includegraphics[width=3.25in]{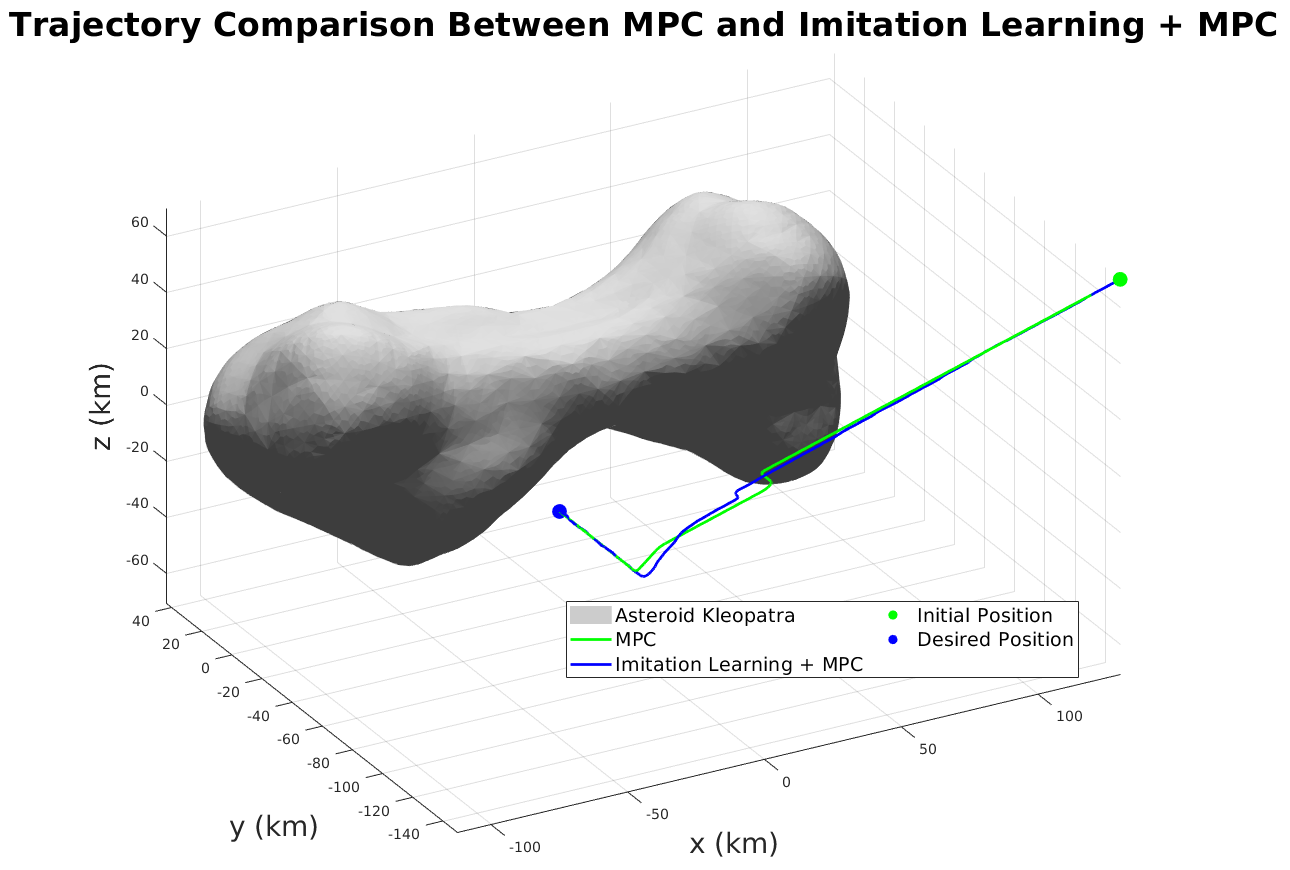}
\caption{Trajectory comparison of hybrid and MPC controller on a test run.}
\label{fig:FreeViewTrajectories}
\end{figure}

A modified version of the script used for data generation with the MPC was then used for forward testing of the trained model. Sensor data was sent from Gazebo back to MATLAB where it was kept in a buffer and fed to the trained imitation learning model for predicting optimal thrust commands. After limiting the thrust commands generated by the model to the same limits as the thruster had during the MPC runs, the generated commands were applied to the spacecraft. Observation of the model's performance in this capacity yielded some important results which would influence the development and deployment of a third controller which took a hybrid approach utilizing both the MPC and the trained model\cite{Sinha2023-st,Hobbs2021-kh}. This third controller is the aforementioned hybrid MPC guided imitation learning controller, and its performance compared to the MPC's on a test trajectory can be seen in Figure~\ref{fig:FreeViewTrajectories}. The optimality of the MPC's trajectory seen in Figure~\ref{fig:FreeViewTrajectories} resulted from a weighted minimization of state error and control expenditure.

While the trained model output similar thrust commands to the MPC the majority of the time in testing, as was predicted by the correlation analysis of the test data, the model's performance when forward tested was poor, with the model failing to reach the desired endpoint and crashing into the asteroid. This was hypothesized to stem from a few issues with the approach taken when creating the model. First, since the model is deterministic and the states of the system tend to change slowly, entering a region in the state space where the model performed poorly would result in a stream of bad control commands being generated which had the cumulative potential to do great damage to the trajectory. Second, as errors were made and the trajectory deviated from the optimal trajectory, the model's performance degraded further as it faced situations increasingly different from what it was trained on. While the attempts made to minimize the impact of covariate shift and improve generalizability did help improve forward performance somewhat, this showed that it remains a somewhat stubborn and fundamental problem which likely requires an even larger training dataset to fully mitigate. Third, the model had no constraints placed on where it could go in terms of proximity to the asteroid, and so it could easily fly into regions the MPC would've never reached (the MPC was constrained to not take trajectories within an ellipsoid encompassing Kleopatra to avoid crashing into it). In those regions close to the asteroid where good performance was most critical, the model was again faced with situations it had never experienced, often with disastrous results. That being said, its important to keep in mind that comparisons to the MPC's forward performance are a bit unfair to the imitation learning based controller as the MPC was given its exact state rather than sensor data and so could naturally exactly navigate to the desired end state, as opposed to the imitation learning based controller which had to deal with sensor noise as well as handling state estimation internally.

In order to improve the performance of the model while still retaining most of the computational cost benefits, an alternative hybrid approach was also developed which generated MPC commands every 1 in 10 time steps of the simulation. This controller could be deployed in situations where an exact state is occasionally available, or in situations where the exact state is available but the energy and computational savings associated with using trained models over optimization algorithms are desired. The optimal commands occasionally generated by the MPC in this controller were compared to the predicted commands to check if the trained model was currently viable for use. In order to perform this comparison the following inequalities were used: 

\begin{equation}
\begin{aligned}
        & \left\| \mathbf{u_{MPC}} - \mathbf{u_{model}} \right\| < \varepsilon_{magnitude} \\
        & \quad\quad\quad\quad\quad\quad\quad\quad \mathbf{OR} \\
        & \arctan\left(\frac{\mathbf{u_{MPC}} \cdot \mathbf{u_{model}}}{\left\| \mathbf{u_{MPC}} \times \mathbf{u_{model}} \right\|} \right)  < \varepsilon_{angle}
\end{aligned}
\end{equation}

If either condition was fulfilled, with one measuring output magnitude error and the other commanded thrust direction error, the model was considered viable and MPC solutions wouldn't be generated until the next check was run, saving energy and computation time. If instead the model wasn't considered to be viable by these metrics, then the controller would revert back to the MPC for all time steps until the next check was run. 

\section{Results}

\begin{figure*}
\centering
\includegraphics[width=5in]{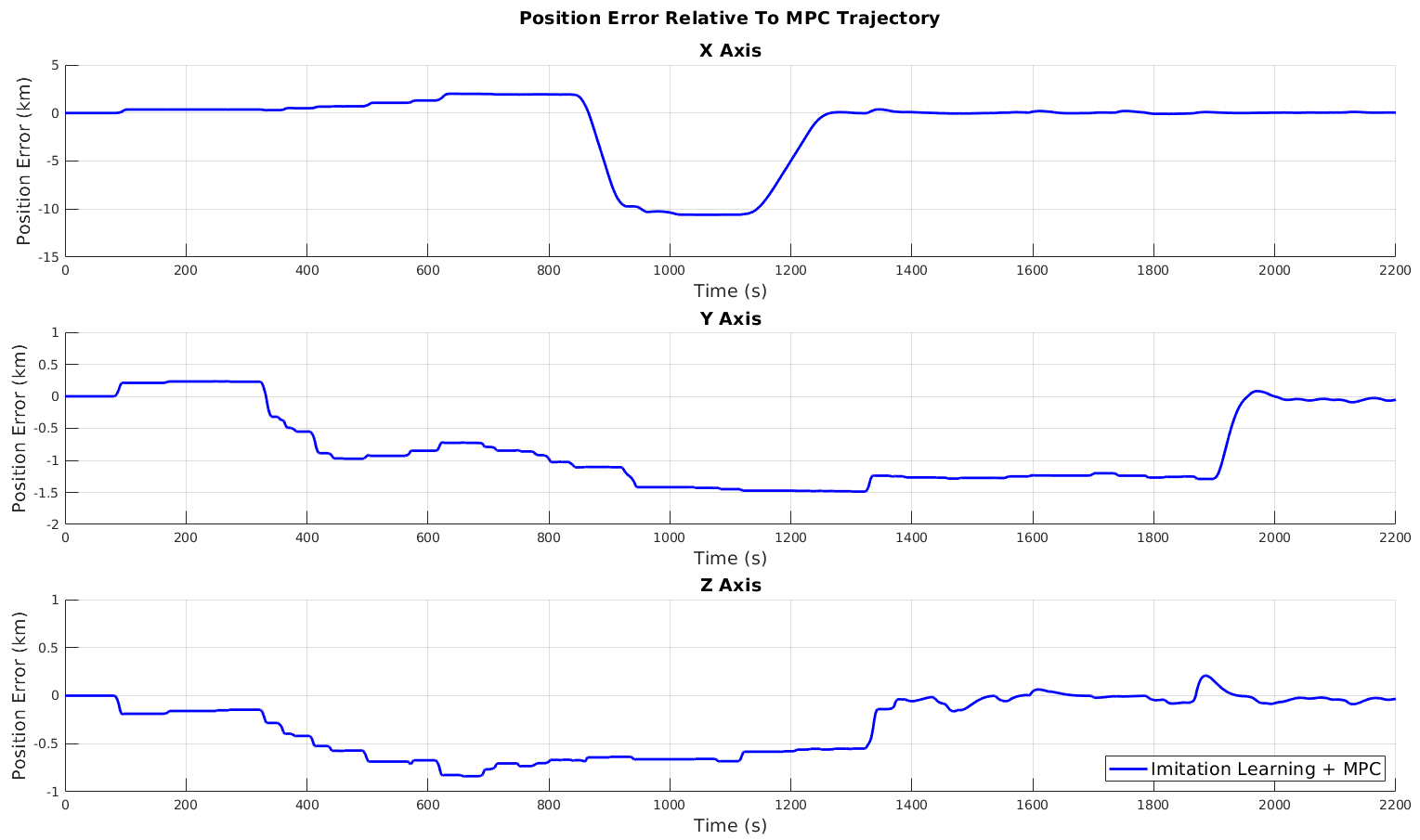}
\caption{MPC guided imitation learning controller position error over time relative to MPC trajectory}
\label{fig:LargePosError}
\end{figure*}

As was mentioned previously, results from the imitation learning controller were somewhat disappointing, with it failing to reach the exact desired point and occasionally crashing into the asteroid during testing. However, after noting the deficiencies and attempting to correct them with the creation of the hybrid MPC guided imitation learning controller, performance significantly improved. In Figures~\ref{fig:FreeViewTrajectories} and~\ref{fig:LargePosError}, the hybrid MPC guided imitation learning controller can be seen to almost perfectly match the base MPC trajectory. When given the same starting state, the purely imitation learning based controller only made small initial progress along the MPC trajectory before deviating and eventually flying out of control as its mistakes compounded into further mistakes and it soon reached the limits of the range of its lidar sensors for detecting Kleopatra, making further attempts at accurate control hopeless.

\begin{figure*}
    \centering
    \includegraphics[width=5in]{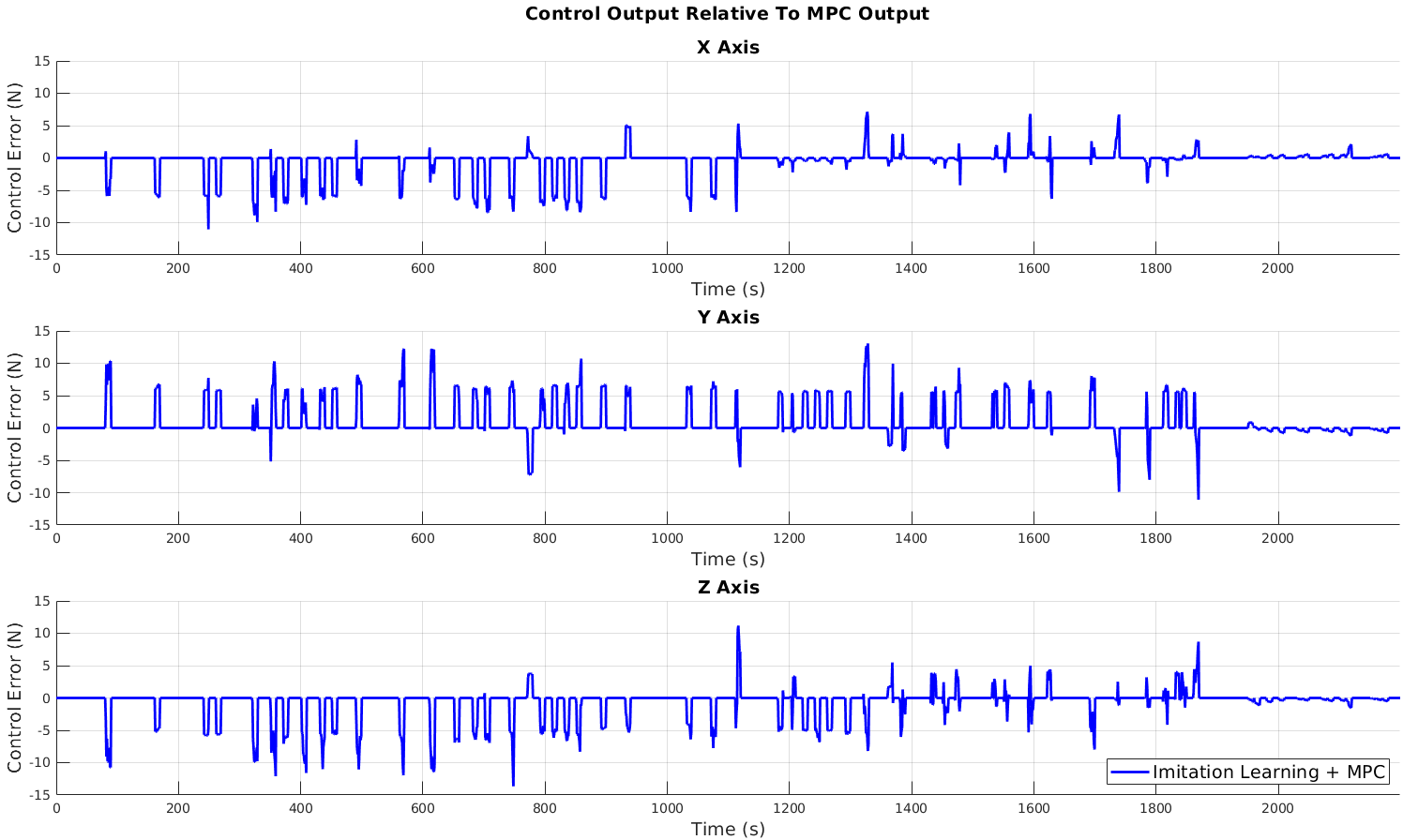}
    \caption{Control effort comparison of MPC guided imitation learning controller relative to MPC control output}
    \label{fig:ControlErrorComparisons}
\end{figure*}

The hybrid MPC guided imitation learning controller was able to use the learned model up to 70\% of the time in testing while reliably reaching the desired end state. This result, combined with the observation of inference times dropping from $\sim$0.473 seconds to $\sim$0.053 seconds when using the trained model instead of the MPC, demonstrates that the developed hybrid controller remains effective for control while offering significant computational efficiency gains over traditional methods. For reference, the MPC solutions were generated on a CPU with a thermal design power (TDP) of 155 W while inference for the trained model was performed on a GPU with a TDP of 300 W. This means that when using the trained model, we can expect to use only $\sim$21.6\% of the energy used by the MPC, a significant achievement.

It is important to note, however, that once developed, the model in this paper does not require high-power GPU requirements. The GPU is used to accelerate neural network training for its dedicated processing capabilities, but for the forward implementation and deployment of the model, the calculations can be performed with conventional hardware that is currently used in spacecraft. Hence, the model was tested on the same CPU as the MPC and achieved an inference time of $\sim$0.138 seconds, resulting in only $\sim$29.2\% of the energy consumption as compared to the MPC. This increase in computational efficiency, even on hardware that could be rated for space applications demonstrates the ability of this model to be a realistic solution. 

Our method for retrieving the lidar scans, using the 12 planar sensors, is purely a consequence of the integration with ROS and the limitation in our hardware speed, as running more than 12 vertical scans would result is significant degradation in our computer speed. Comparatively, state-of-the-art lidar technology onboard spacecraft can sample at rates up to 4000 Hz and acquire up to 25000 points/second \cite{lidar-ommatidia}. These data sampling rates and fidelity of points that can be achieved via real-life lidar systems far exceeds the capabilities of our lidar simulations, hence showing that with training on real systems, further improvement is possible \cite{survey-lidar,acp-11-7045-2011}. 

In Figure~\ref{fig:ControlErrorComparisons}, which compares the developed MPC guided imitation learning controller's output to what the MPC would output if it was on the same trajectory, it can be seen that the hybrid controller spent most of its time in agreement with the MPC, while still receiving occasional corrective actions from the MPC to maintain compliance with the MPC's constraints. Its important to note that small violations from the imitation learning controller on the constraints on velocity and positioning placed on the MPC, resulted in large disagreements in control outputs, making this metric highly sensitive to small changes/errors in spacecraft state. The MPC guided imitation learning controller, owing to occasional MPC intervention, remained constrained to a set of states where the imitation learning model was able to exhibit a high level of performance as it didn't have to face scenarios too dissimilar to what it was trained on. The spikes in control error seen for the MPC guided imitation learning controller for the most part point to situations where a corrective thrust was deemed necessary by the MPC and the validation check of the outputs of the imitation learning model against the MPC hadn't yet been performed. Once these corrective measures were taken, the error again dropped to near zero levels. As the MPC guided imitation learning controller approached the desired final state, it's error dropped to consistent small values, suggesting that the imitation learning model can successfully recognize when it has reached the target states. In the future, the optimality of the trained controllers outputs in general cases could likely be further improved by training on even larger datasets and providing examples of correcting for overshoots of the desired end point as model performance seemed to significantly degrade after overshooting the target point, likely due to a lack of examples of overshoots in the training data.

\section{Conclusions}

The usage of imitation learning techniques to train models presents many potential benefits over the traditional GNC pipelines in use today. In situations where dynamics are uncertain and navigation is difficult, trained agents can offer suitable replacements for traditional control methods when used with some sort of runtime assurance\cite{Feron2022-bd,Kalaria2023-su,Sinha2023-st}. These agents can also offer the advantages of lessened computational/heat/power requirements when compared to traditional optimal controllers. That being said, the performance of these trained agents, by virtue of them being trained to simply imitate the expert agent as best as possible, can be expected to be reduced when compared to the expert agents themselves. With more extensive training on diverse datasets perhaps parity can be achieved between trained and expert models, but for now, it seems that some performance must be sacrificed to experience the benefits that imitation learning and ML based controllers have to offer.

\bibliographystyle{IEEEtran}
\bibliography{ref}

\begin{thebibliography}{10}
\providecommand{\url}[1]{#1}
\csname url@samestyle\endcsname
\providecommand{\newblock}{\relax}
\providecommand{\bibinfo}[2]{#2}
\providecommand{\BIBentrySTDinterwordspacing}{\spaceskip=0pt\relax}
\providecommand{\BIBentryALTinterwordstretchfactor}{4}
\providecommand{\BIBentryALTinterwordspacing}{\spaceskip=\fontdimen2\font plus
\BIBentryALTinterwordstretchfactor\fontdimen3\font minus \fontdimen4\font\relax}
\providecommand{\BIBforeignlanguage}[2]{{%
\expandafter\ifx\csname l@#1\endcsname\relax
\typeout{** WARNING: IEEEtran.bst: No hyphenation pattern has been}%
\typeout{** loaded for the language `#1'. Using the pattern for}%
\typeout{** the default language instead.}%
\else
\language=\csname l@#1\endcsname
\fi
#2}}
\providecommand{\BIBdecl}{\relax}
\BIBdecl

\bibitem{NewRef1}
G.~M. Nehma, M.~Tiwari, and M.~Lingam, ``Advancements in spacecraft rendezvous: Leveraging koopman theory over clohessy-wiltshire equations,'' p. 1943, 2025.

\bibitem{NewRef2}
E.~Whitney, B.~P.~R. Gopu, and M.~Tiwari, ``{3D} global localization of a {UAV} using {2D} monte carlo localization and ground plane extraction,'' p. 1536, 2025.

\bibitem{NewRef3}
T.~Mahendrakar, R.~T. White, and M.~Tiwari, ``{SpY}: A context-based approach to spacecraft component detection,'' \emph{arXiv [cs.CV]}, Jun. 2024.

\bibitem{NewRef4}
G.~Nehma and M.~Tiwari, ``Leveraging kans for enhanced deep koopman operator discovery,'' \emph{arXiv [cs.LG]}, 2024.

\bibitem{NewRef5}
\BIBentryALTinterwordspacing
T.~Mahendrakar, R.~T. White, M.~Tiwari, and M.~Wilde, ``\BIBforeignlanguage{en}{Unknown non-cooperative spacecraft characterization with lightweight convolutional neural networks},'' \emph{\BIBforeignlanguage{en}{J. Aerosp. Comput. Inf. Commun.}}, pp. 1--6, Mar. 2024. [Online]. Available: \url{https://arc.aiaa.org/doi/10.2514/1.I011343}
\BIBentrySTDinterwordspacing

\bibitem{NewRef6}
G.~Nehma, M.~Tiwari, and M.~Lingam, ``Deep learning based dynamics identification and linearization of orbital problems using koopman theory,'' \emph{arXiv [math-ph]}, 2024.

\bibitem{NewRef7}
M.~Tiwari, G.~Nehma, and B.~Lusch, ``Computationally efficient data-driven discovery and linear representation of nonlinear systems for control,'' \emph{IEEE Control Syst. Lett.}, vol.~7, pp. 3373--3378, Sep. 2023.

\bibitem{NewRef8}
D.~Zuehlke, M.~Tiwari, K.~Jebari, and K.~B. Kidambi, ``Rendezvous and proximity operations in cislunar space using linearized dynamics for estimation,'' \emph{Aerospace}, vol.~10, no.~8, p. 674, Jul. 2023.

\bibitem{NewRef9}
T.~Mahendrakar, R.~T. White, M.~Wilde, and M.~Tiwari, ``Spaceyolo: A human-inspired model for real-time, on-board spacecraft feature detection,'' pp. 01--11, Mar. 2023.

\bibitem{NewRef10}
M.~Tiwari, R.~Prazenica, and T.~Henderson, ``\BIBforeignlanguage{en}{Direct adaptive control of spacecraft near asteroids},'' \emph{\BIBforeignlanguage{en}{Acta Astronaut.}}, vol. 202, pp. 197--213, Jan. 2023.

\bibitem{NewRef111}
\BIBentryALTinterwordspacing
C.~W. Hays, M.~Tiwari, T.~Henderson, and R.~J. Prazenica, ``A geometric mechanics approach to direct adaptive model predictive control,'' in \emph{AIAA SCITECH 2022 Forum}, ser. AIAA SciTech Forum.\hskip 1em plus 0.5em minus 0.4em\relax American Institute of Aeronautics and Astronautics, Dec. 2021. [Online]. Available: \url{https://doi.org/10.2514/6.2022-2380}
\BIBentrySTDinterwordspacing

\bibitem{Li2023-if}
S.~Li, H.-T. Nguyen, and C.~C. Cheah, ``A theoretical framework for {End-to-End} learning of deep neural networks with applications to robotics,'' \emph{IEEE Access}, vol.~11, pp. 21\,992--22\,006, 2023.

\bibitem{Kalaria2023-su}
D.~Kalaria, Q.~Lin, and J.~M. Dolan, ``Towards safety assured {End-to-End} {Vision-Based} control for autonomous racing,'' Mar. 2023.

\bibitem{Scaramuzza2022-le}
D.~Scaramuzza and E.~Kaufmann, ``Learning agile, vision-based drone flight: From simulation to reality,'' \emph{The International Symposium of Robotics}, 2022.

\bibitem{Tiwari2022-jt}
M.~Tiwari, R.~Prazenica, and T.~Henderson, ``Direct adaptive control of spacecraft near asteroids,'' \emph{Acta Astronaut.}, Oct. 2022.

\bibitem{Tiwari2022-ki}
M.~Tiwari, E.~Coyle, and R.~J. Prazenica, ``\BIBforeignlanguage{en}{{Direct-Adaptive} nonlinear {MPC} for spacecraft near asteroids},'' \emph{\BIBforeignlanguage{en}{Aerospace}}, vol.~9, no.~3, p. 159, Mar. 2022.

\bibitem{Attia2018-cy}
A.~Attia and S.~Dayan, ``Global overview of imitation learning,'' Jan. 2018.

\bibitem{Zare2023-kf}
M.~Zare, P.~M. Kebria, A.~Khosravi, and S.~Nahavandi, ``A survey of imitation learning: Algorithms, recent developments, and challenges,'' Sep. 2023.

\bibitem{Lee2018-qh}
K.~Lee, K.~Saigol, and E.~A. Theodorou, ``Safe end-to-end imitation learning for model predictive control,'' Mar. 2018.

\bibitem{Pan2020-rt}
Y.~Pan, C.-A. Cheng, K.~Saigol, K.~Lee, X.~Yan, E.~A. Theodorou, and B.~Boots, ``\BIBforeignlanguage{en}{Imitation learning for agile autonomous driving},'' \emph{\BIBforeignlanguage{en}{Int. J. Rob. Res.}}, vol.~39, no. 2-3, pp. 286--302, Mar. 2020.

\bibitem{Chen2020-ab}
D.~Chen, B.~Zhou, V.~Koltun, and P.~Kr{\"a}henb{\"u}hl, ``Learning by cheating,'' in \emph{Proceedings of the Conference on Robot Learning}, ser. Proceedings of Machine Learning Research, L.~P. Kaelbling, D.~Kragic, and K.~Sugiura, Eds., vol. 100.\hskip 1em plus 0.5em minus 0.4em\relax PMLR, 2020, pp. 66--75.

\bibitem{Zhang2016-ip}
J.~Zhang and K.~Cho, ``{Query-Efficient} imitation learning for {End-to-End} autonomous driving,'' May 2016.

\bibitem{Alexander2024-my}
P.~Alexander, D.~Alperen, W.~David, H.~Shashank, O.~Ryan, K.~Alexey, D.~Bertrand, N.~David, M.~Urs, B.~Ruchi, B.~Stan, and S.~Nikolai, ``Mitigating covariate shift in imitation learning for autonomous vehicles using latent space generative world models,'' \emph{arXiv [cs.RO]}, Sep. 2024.

\bibitem{Codevilla2018-sq}
F.~Codevilla, M.~Muller, A.~Lopez, V.~Koltun, and A.~Dosovitskiy, ``End-to-end driving via conditional imitation learning,'' in \emph{2018 IEEE International Conference on Robotics and Automation (ICRA)}.\hskip 1em plus 0.5em minus 0.4em\relax IEEE, May 2018, pp. 4693--4700.

\bibitem{Kaufmann2020-mr}
E.~Kaufmann, A.~Loquercio, R.~Ranftl, M.~M{\"u}ller, and {others}, ``Deep drone acrobatics,'' \emph{arXiv preprint arXiv}, 2020.

\bibitem{Sinha2023-st}
R.~Sinha, E.~Schmerling, and M.~Pavone, ``Closing the loop on runtime monitors with {Fallback-Safe} {MPC},'' in \emph{2023 62nd {IEEE} Conference on Decision and Control ({CDC})}.\hskip 1em plus 0.5em minus 0.4em\relax IEEE, Dec. 2023, pp. 6533--6540.

\bibitem{Hobbs2021-kh}
K.~Hobbs, M.~Mote, M.~Abate, S.~Coogan, and E.~Feron, ``Run time assurance for {Safety-Critical} systems: An introduction to safety filtering approaches for complex control systems,'' Oct. 2021.

\bibitem{lidar-ommatidia}
O.~L. S.L, ``Q2 laser radar,'' 2024, \url{https://ommatidia-lidar.com/product/q2/} [Accessed: 12/25/2024].

\bibitem{survey-lidar}
\BIBentryALTinterwordspacing
J.~A. Christian and S.~Cryan, \emph{A Survey of LIDAR Technology and its Use in Spacecraft Relative Navigation}. [Online]. Available: \url{https://arc.aiaa.org/doi/abs/10.2514/6.2013-4641}
\BIBentrySTDinterwordspacing

\bibitem{acp-11-7045-2011}
\BIBentryALTinterwordspacing
K.~Knobelspiesse, B.~Cairns, M.~Ottaviani, R.~Ferrare, J.~Hair, C.~Hostetler, M.~Obland, R.~Rogers, J.~Redemann, Y.~Shinozuka, A.~Clarke, S.~Freitag, S.~Howell, V.~Kapustin, and C.~McNaughton, ``Combined retrievals of boreal forest fire aerosol properties with a polarimeter and lidar,'' \emph{Atmospheric Chemistry and Physics}, vol.~11, no.~14, pp. 7045--7067, 2011. [Online]. Available: \url{https://acp.copernicus.org/articles/11/7045/2011/}
\BIBentrySTDinterwordspacing

\bibitem{Feron2022-bd}
E.~Feron, O.~Sanni, M.~L. Mote, D.~Delahaye, T.~Khamvilai, M.~Gariel, and S.~Saber, ``Ariadne: A common-sense thread for enabling provable safety in air mobility systems with unreliable components,'' \emph{AIAA SCITECH 2022 Forum}, Jan. 2022.

\end{thebibliography}

\thebiography
\begin{biographywithpic}
{Patrick Quinn}{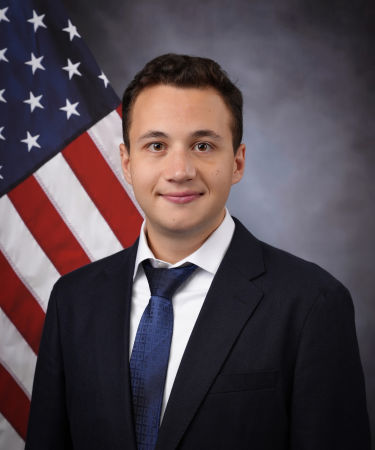}
is a master's student at Florida Institute of Technology currently studying the application of neural networks for end-to-end control systems. In the past, he has worked on research relating to optimal control and reinforcement learning for spacecraft and quadcopters, as well as integrating experimental sensors with open-source drone software.
\end{biographywithpic} 

\begin{biographywithpic}
{George Nehma}{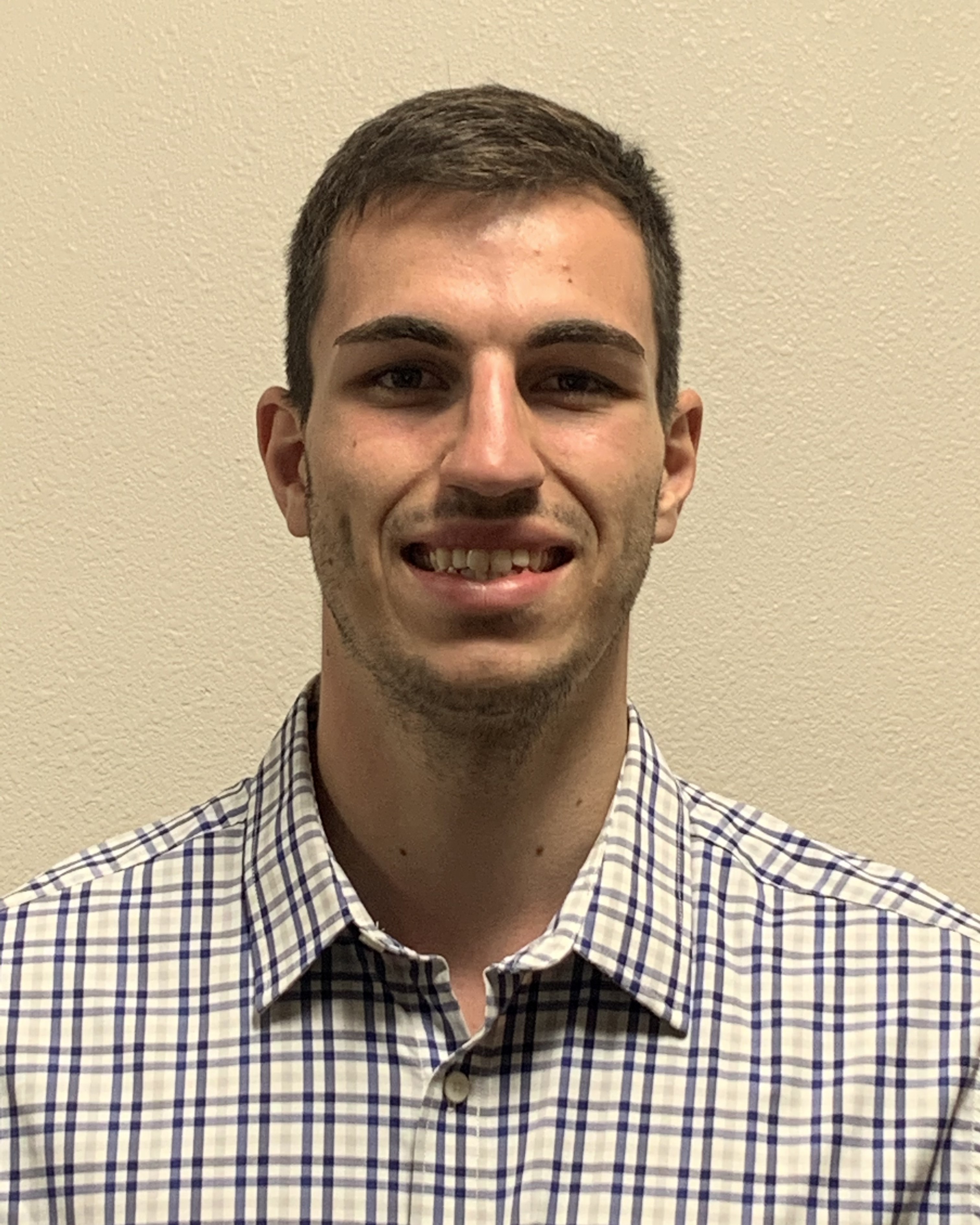}
is a PhD student at Florida Institute of Technology currently studying the application of neural networks for end-to-end control systems. In the past, he has worked on research relating to the application of Neural Koopman Theory for classical control systems in orbital dynamics and robotics as well as adaptive controls for spacecraft attitude controls.
\end{biographywithpic} 

\begin{biographywithpic}
{Dr. Madhur Tiwari}{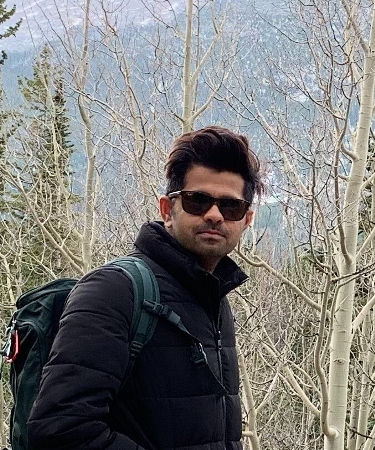}
is the Director of the Autonomy Lab and Assistant Professor of Aerospace Engineering at Florida Institute of Technology. He specializes in robotics, machine learning and control for aerospace systems. Currently, he is teaching Spaceflight Mechanics and Modern Control Theory at Florida Institute of Technology.

\end{biographywithpic}

\end{document}